\documentclass[10pt,twocolumn,letterpaper]{article}
\pdfoutput=1

\usepackage{cvpr}
\usepackage{times}
\usepackage{epsfig}
\usepackage{graphicx}
\usepackage{amsmath}
\usepackage{amssymb}
\usepackage{booktabs}
\usepackage{multirow}
\usepackage{tabularx}

\newcolumntype{L}[1]{>{\raggedright\arraybackslash}p{#1}}
\newcolumntype{C}[1]{>{\centering\arraybackslash}p{#1}}
\newcolumntype{R}[1]{>{\raggedleft\arraybackslash}p{#1}}

\usepackage[caption=false,font=footnotesize]{subfig}

\newcommand{\deflen}[2]{%      
    \expandafter\newlength\csname #1\endcsname
    \expandafter\setlength\csname #1\endcsname{#2}%
}

\usepackage[pagebackref=true,breaklinks=true,letterpaper=true,colorlinks,bookmarks=false]{hyperref}

\cvprfinalcopy % *** Uncomment this line for the final submission

\def\cvprPaperID{2699} % *** Enter the CVPR Paper ID here

\ifcvprfinal\fi
\begin{document}

%%%%%%%%% TITLE
\title{Deep Surface Normal Estimation with Hierarchical RGB-D Fusion}

\author{Jin Zeng$^1$ \quad Yanfeng Tong$^{1,2*}$ \quad Yunmu Huang$^{1}$\thanks{ indicates equal contribution.} \quad Qiong Yan$^1$ \quad Wenxiu Sun$^1$ \\
Jing Chen$^2$ \quad Yongtian Wang$^2$\\
$^1$SenseTime Research \quad \quad \quad \quad $^2$Beijing Institute of Technology\\
{\tt\small $^1$\{zengjin, tongyanfeng, huangyunmu, yanqiong, sunwenxiu\}@sensetime.com}
% \and
% First line of institution2 address
\\
{\tt\small $^2$\{chen74jing29, wyt\}@bit.edu.cn}
\\
{\tt\small https://github.com/jzengust/RGBD2Normal}
}

\maketitle
\thispagestyle{empty}

%%%%%%%%% ABSTRACT
\begin{abstract}
   The growing availability of commodity RGB-D cameras has boosted the applications in the field of scene understanding. However, as a fundamental scene understanding task, surface normal estimation from RGB-D data lacks thorough investigation.
   In this paper, a hierarchical fusion network with adaptive feature re-weighting is proposed for surface normal estimation from a single RGB-D image. 
   Specifically, the features from color image and depth are successively integrated at multiple scales to ensure global surface smoothness while preserving visually salient details. Meanwhile, the depth features are re-weighted with a confidence map estimated from depth before merging into the color branch to avoid artifacts caused by input depth corruption.
   Additionally, a hybrid multi-scale loss function is designed to learn accurate normal estimation given noisy ground-truth dataset. 
   Extensive experimental results validate the effectiveness of the fusion strategy and the loss design, outperforming state-of-the-art normal estimation schemes.
\end{abstract}

%%%%%%%%% BODY TEXT
\section{Introduction}
\label{sec:intro}
\begin{figure}[t]
\centering
    \subfloat[RGB Image]{\includegraphics[width=\realwidthIntro]{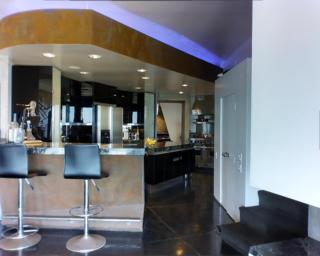}}\hspace{\gauinterIntro}
    \subfloat[Sensor Depth]{\includegraphics[width=\realwidthIntro]{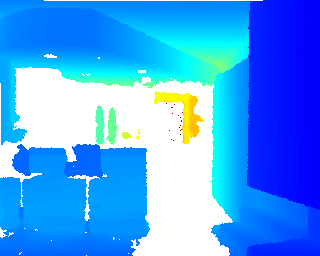}}\hspace{\gauinterIntro}
    \subfloat[Using RGB]{\includegraphics[width=\realwidthIntro]{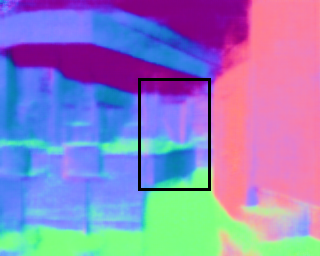}}\hspace{\gauinterIntro}
    \subfloat[Using Depth]{\includegraphics[width=\realwidthIntro]{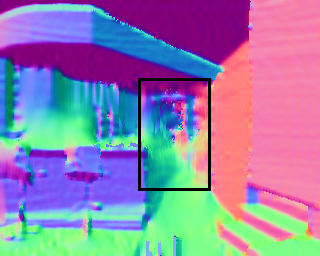}}\hspace{\gauinterIntro}
    \subfloat[Early Fusion]{\includegraphics[width=\realwidthIntro]{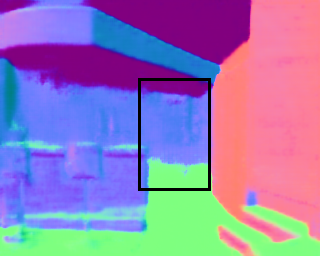}}\hspace{\gauinterIntro}
    \subfloat[Hierarchical Fusion]{\includegraphics[width=\realwidthIntro]{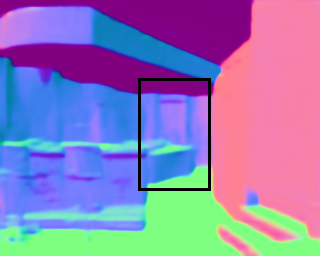}}       
 \caption{Example in Matterport3D dataset. (a) RGB input; (b) depth input; normal estimation with (c) single RGB \cite{zhang2017physically}, (d) depth inpainting \cite{fouhey2013data}, (e) RGB-D early fusion \cite{zhang2018deep}, (f) proposed hierarchical RGB-D fusion.}
 \label{fig:source}
\end{figure}

% RGB
Per-pixel surface normal estimation has been extensively studied in the recent years.
% achieving satisfactory results thanks to the use of convolutional neural network. 
Previous works on normal estimation mostly assume single RGB image input \cite{eigen2015predicting,wang2016surge,bansal2016marr,zhang2017physically}, providing satisfying results in most cases despite loss of shape features and erroneous results at the highlight or dark areas, as shown in Fig.\,\ref{fig:source}(c).

%D
RGB-D cameras are now commercially available, leading to a great performance enhancement in the applications of scene understanding, \eg, semantic segmentation \cite{wang2018depthconv,chu2018surfconv,su2018splatnet}, object detection \cite{gupta2014learning,Qi2018CVPR}, 3D reconstruction \cite{Litany2018CVPR,newcombe2015dynamicfusion,izadi2011kinectfusion}, \etc. With the depth given by sensors, normal can be easily calculated via a least square optimization \cite{qi2018geonet,fouhey2013data} as used in the widely used NYUv2 dataset \cite{silberman2012indoor}, but the quality of the normal suffers from the corruption in depth, \eg, sensor noise along object edges or missing pixels due to glossy, black, transparent, and distant surfaces \cite{Su_2018_CVPR}, as shown in Fig.\,\ref{sec:intro}(d).

\deflen{gauinterIntro}{0pt}
\deflen{gauwidthIntro}{82pt}
\deflen{realwidthIntro}{75pt}

% Motivation: lack of fusion method
This motivates us to combine the advantages of color and depth inputs while compensating for the deficiency of each other in the task of normal estimation.
Specifically, the RGB information is utilized to fill the missing pixels in depth; meanwhile the depth clue is merged into RGB results to enhance sharp edges and correct erroneous estimation, resulting in a complete normal map with fine details. 
However, research on combining RGB and depth for normal estimation has not been extensively studied. To the best of our knowledge, the only work considering RGB-D input for normal estimation adopts \textit{early fusion}, \textit{i.e.}, using depth as an additional channel to the RGB input, leading to little performance improvement compared with the methods using the RGB input only \cite{zhang2018deep}. The lack of proper network design for combining the geometric information in depth and color image is an impediment to fully take advantage of the depth sensor.

% Fusion problem: 
% early fusion: cannot handle well signals from different domains; late fusion does not fully utilize depth, output from RGB branch can be too different from that of depth branch in large scale and difficult to rectify the error without affecting edges
% hierarchical fusion to fully utilize depth feature in refining RGB in different scales: small scales for overall surface orientation fidelity, large scale for detail refinement and edge preservation
% and estimate confidence map to adaptively reweight depth feature to ensure smooth transition at rectified areas
% Ours
Different from previous works on normal estimation with RGB-D using early fusion \cite{zhang2018deep}, we propose to merge the features from RGB and depth branches at multiple scales at the decoder side in a hierarchical manner, in order to guarantee both global surface smoothness and local sharp features in the fusion results. 
Additionally, a pixel-wise confidence map is estimated from the depth input for re-weighting depth features before merging into RGB branch, so as to reduce artifact from depth with a smaller confidence on missing pixels and those along the object edges. An example is shown in Fig.\,\ref{fig:source}, where the proposed scheme outperforms state-of-the-art RGB-based, depth-based, RGBD-based methods.

% Challenge II: dataset
Apart from the lack of RGB-D fusion schemes, the shortage of datasets providing sensor depth and ground-truth depth pairs is another obstacle for RGB-D normal estimation since the performance of DNN approaches is affected by the dataset quality \cite{Pang_2018_CVPR,zeng2018deep}.
The widely used training datasets for normal estimation, \eg, NYUv2 \cite{silberman2012indoor}, do not provide complete ground-truth normal for the captured RGB-D images since it is directly computed from the captured depth after inpainting \cite{levin2004colorization}. If trained on NYUv2, the network is up to approximate an inpainting algorithm. 

% Dataset problem
% How to solve: hybrid loss function, small scale with l2 loss to suppress large error, large scale with l1 loss to avoid noise along edges based on ground-truth noise analysis
Instead we use Matterport3D \cite{Matterport3D} and ScanNet \cite{dai2017scannet} datasets with RGB-D captured by camera and ground-truth normal obtained via multiview reconstruction provided by \cite{zhang2018deep}.
Nevertheless, the ground-truth is not perfect due to the multiview reconstruction error, especially at object edges which is crucially for visual evaluation. To overcome the artifact in the ground-truth, we propose a hybrid multi-scale loss function based on the noise statistics in the ground-truth normal map, using $L_1$ loss at the large resolution to obtain sharper results, and $L_2$ loss at small resolution to ensure coarse scale accuracy.

% contribution
In summary, the main contributions of our work are:
\begin{itemize}
    \item By incorporating RGB and depth inputs via the proposed hierarchical fusion scheme, the two inputs are able to complement each other in the normal estimation, refine details with depth, and fill the missing depth pixels with color;
    \item With the confidence map for depth feature re-weighting, the effect of artifacts in the depth features is reduced;
    \item A hybrid multi-scale loss function is designed by analyzing the noise statistics in the ground-truth, providing sharp results with high fidelity despite the imperfect ground-truth.
\end{itemize}
Comparison with the state-of-the-art approaches and extensive ablation study validates the design of network structure and loss function. The paper is organized as follows. Related works are discussed in Section\,\ref{sec:related}, and Section\,\ref{sec:method} provides a detailed discussion of the proposed method. Ablation study and comparison with state-of-the-art methods are demonstrated in Section\,\ref{sec:exp} and the work is concluded in Section\,\ref{sec:con}.

%-------------------------------------------------------------------------

\section{Related Work}
\label{sec:related}
\begin{figure*}[t]
\centering
    \includegraphics[width=500pt]{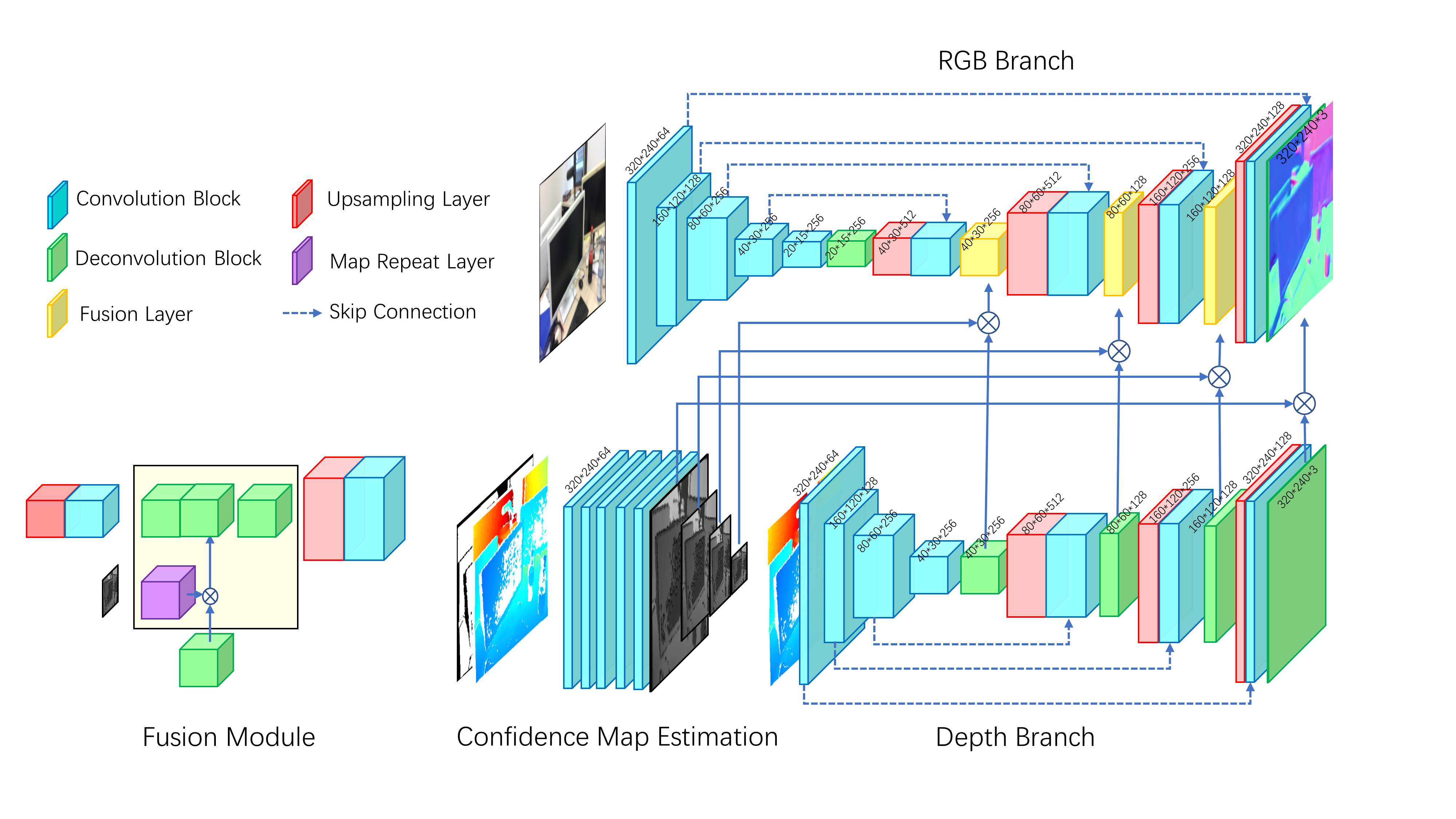}
 \caption{Proposed hierarchical RGB-D fusion composed of RGB branch at the upper side, depth branch at the lower-right side, confidence map module at the lower-left side. The fusion module is abstracted as a fusion layer in the fusion network and illustrated at the lower-left side. An input with size $320\times 240$ is used for demonstration.}
 \label{fig:framework}
\end{figure*}

\subsection{Surface Normal Estimation}
\textbf{RGB-based}
Previous works mostly used a single RGB image as input.
Eigen \etal \cite{eigen2015predicting} designed a three-scale convolution network architecture that produced a coarse global prediction with full image first and then refined it with local finer-scale network. 
Wang \etal \cite{wang2015designing} proposed a network structure that integrated different geometric information like local, global, and vanishing point information to predict the surface normal. 
% Considering the details of edge are relatively poor, Liu \etal \cite{Liu2016Learning} add a continuous conditional random field (CRF) on the top of CNN to smooth super-pixel-based prediction. 
More recently, Bansal \etal \cite{bansal2016marr} proposed a skip-connected structure to concatenate the CNN response at different scales to capture corresponding details at each scale, and Zhang \etal \cite{zhang2017physically} adopted a U-Net structure and achieved state-of-the-art performance.

Due to the difficulty in extracting geometric information and texture interference from the RGB input, the details of predictions are poor, with wrong results in the area of insufficient lighting or high lighting.

\textbf{Depth-based}
Surface normal can be inferred from depth with geometric method, which depends on the neighboring pixels' relative depth geometrically \cite{zhang2018deep}.
%But in common dataset , \eg, NYU  \cite{silberman2012indoor}, Matterport3D   \cite{Chang2017Matterport3D}, ScanNet  \cite{dai2017scannet} use depth camera to capture depth image. 
However, the depth camera used in common datasets, \eg, NYUv2  \cite{silberman2012indoor}, Matterport3D \cite{Matterport3D}, ScanNet  \cite{dai2017scannet} often fails to sense the depth on glossy, bright, transparent and faraway surfaces \cite{zhang2018deep,zeng20183d}, resulting in holes and corruptions in the obtained depth images. 
To overcome missing pixels in normal map inferred from depth, some works proposed to inpaint depth images using RGB images \cite{Daniel2013Depth,Gong2013Guided,Liu2012Guided,Thabet20143D,Zhang2017Probability}. 
Silberman \etal \cite{silberman2012indoor} used optimization-based method \cite{levin2004colorization} to fill the holes in depth maps. Zhang \etal \cite{zhang2018deep} used a convolutional network to predict pixel-wise surface normal with a single RGB image, then used the predicted normal to fill holes in raw depth.

Nevertheless, depth inpainting cannot handle large holes in depth; also, the noise in depth will undermine depth-based normal estimation performance.

\textbf{Normal-depth consistency based}
There is a strong geometric correlation between the depth and the surface normal. 
Normal can be calculated from the depth of neighboring pixels, and depth can be refined with normal variation. 
For example, Wang \etal \cite{wang2016surge} proposed a four-stream convolutional neural network to detect planar regions, then used a dense conditional random field to smooth results based on depth and surface normal correlation in planar region and planar boundary respectively.
Chen \etal \cite{chen2017surface} established a new dataset, and proposed two loss functions to measure the consistency between predicted normal and depth label for depth and normal prediction.
Qi \etal \cite{qi2018geonet} proposed to predict initial depth and surface normal using color image, then cross-refine each other using geometric consistency.

These methods provide different schemes to promote geometric consistency between normal and depth, but rely on a single RGB input and do not consider noise from depth sensors.

\textbf{RGB-D based}
The RGB-D based normal estimation has not been extensively studied in previous works. Normal estimation with RGB-D input has been briefly discussed in \cite{zhang2018deep} where an early fusion was adopted, reported to be almost the same as using RGB input. However the method is not properly designed and the conclusion is not comprehensive. 
Although 3D reconstruction based methods like \cite{newcombe2015dynamicfusion} can be used in normal estimation, a series of RGB-D images is required for those methods, which is beyond the scope of this paper.
The lack of design in RGB-D fusion for surface normal estimation motivates our work.

\subsection{RGB-D Fusion Schemes}
Despite the lack of study in RGB-D based normal estimation, RGB-D fusion scheme has been explored for other tasks, among which semantic segmentation is the most extensively studied one, \eg, early fusion using RGB-D as a four-channel input \cite{eigen2015predicting}, late fusion \cite{cheng2017localitysensitive}, depth-aware convolution \cite{wang2018depthconv}, or using 3D point cloud format \cite{Qi2018CVPR}.

The difference from those works is that they do not require per-pixel accuracy as much as normal prediction, \textit{i.e.}, the label interior of one object is constant, but for normal estimation, correct prediction at each pixel is required, and the most significant difficulty lies in accurate sharp details. 
Therefore, we adopt hierarchical fusion with confidence map re-weighting to enhance edge preservation in the fusion result without bringing artifacts in depth.

%-------------------------------------------------------------------------
\section{Method}
\label{sec:method}
As illustrated in Fig.\,\ref{fig:framework}, the hierarchical RGB-D fusion network is composed of three modules: RGB branch, depth branch, and confidence map estimation. In this section, we introduce the pipeline for the hierarchical fusion of RGB and depth branches with the fusion module at different scales, and confidence map estimation used inside the fusion module for depth conditioning, after which the hybrid loss function design is detailed.
A detailed architecture of the deep network is provided in the supplementary. 

% The general framework is illustrated in Fig. X, where the color and depth images are first feed into the individual FCN encoders, and then depth features at four difference scales from the decoder side are adaptively re-weighted by confidence map then concatenated to corresponding same-sized color features. A detailed discussion of the hierarchical fusion scheme are provided as follows. 
% (Need figure for demo, and images)

\subsection{Hierarchical RGB-D Fusion}
% Problem formulation
Given color image ${\cal I}_c$ and sensor depth ${\cal I}_d$, we are aimed as estimating surface normal map ${\cal I}_n$ by minimizing its distance from the ground-truth normal ${\cal I}_n^{(\rm gt)}$, \textit{i.e.},
\begin{equation}\label{eq:loss_origin}
\min_{\theta} \quad L({\cal I}_n^{(\rm gt)}, f_\theta({\cal I}_c, {\cal I}_d)),
\end{equation}
where $f_\theta$ denotes the fusion network function to generate normal estimation ${\cal I}_n$ parameterized by the parameters $\theta$, which are end-to-end trained via back propagation. A hierarchical fusion scheme is adopted to merge depth branch into RGB branch for both overall surface orientation rectification and visually salient feature enhancement.

\subsubsection{Network Design}
First, in the RGB branch where the input is the color image ${\cal I}_c$, we adopt a similar network structure as used in \cite{zhang2017physically}, where a fully convolutional network (FCN) \cite{long2015fully} is built with VGG-16 back-bone as illustrated in the RGB branch in Fig.\,\ref{fig:framework}. 
Specifically, the encoder is the same as VGG-16 except that in the last two convolution blocks of the encoder, \textit{i.e.}, conv4 and conv5, the channel number is reduced from 512 to 256 to remove redundant model parameters. The encoder is accompanied with a symmetric decoder, and equipped with skip-connections and shared pooling masks for learning local image features. 

Meanwhile, ${\cal I}_d$ is fed into the depth branch to extract geometric features with a similar network structure as the RGB branch, except that the last convolution block in the RGB encoder is removed to give a simplified model.

The fusion takes place at the decoder side. As shown in Fig.\,\ref{fig:framework}, the depth features (colored in green) at each scale in the decoder are passed into the fusion module and re-weighted with the confidence map (colored in purple) down-sampled and repeated to the same resolution as the depth feature. 
Then the re-weighted depth features are concatenated with the color features with the same resolution and passed through a deconvolution layer to give the fusion output features (colored in yellow). Consequently, the fusion module (denoted as FM for short) at scale $l$ is given as,
\begin{align}\label{eq:map}
    \mathrm{FM}(\mathcal{F}_c^l, \mathcal{F}_d^l|\mathcal{C}^l) = \mathrm{deconv}( \mathcal{F}_c^l \oplus (\mathcal{F}_d^l \odot \mathcal{C}^l)),
\end{align}
where $\mathcal{F}_c^l$, $\mathcal{F}_d^l$ are the features from RGB and depth branches at scale $l$, and $\mathcal{C}^l$ is the confidence map for depth conditioning. $\odot$ denotes element-wise multiplication and $\oplus$ denotes the concatenation operation. The concatenation result after deconvolution layer gives the fusion output.
The fusion is implemented at four scales, where the last scale output gives the final normal estimation. The confidence map estimation is addressed later in Section\,\ref{sec:map}.

\subsubsection{Comparison with Existing RGB-D Fusion Schemes}
Existing RGB-D fusion schemes mostly adopt single-scale fusion. \cite{zhang2018deep} fused RGB-D at the input, \textit{i.e.}, using depth as an additional channel along with RGB. 
However, RGB and depth are from different domains and cannot be properly handled using the same encoder as a four-channel input. 
For example, we adopt the same network structure as in \cite{zhang2017physically}, composed of VGG-16 encoder and a symmetric decoder with skip-connection, and use a RGB-D four-channel input instead of a single RGB to generate the normal as shown in Fig.\,\ref{fig:fusion-loss-compare}(d). The output normal does not exhibit global smoothness, especially in area where depth pixels are missing. This is because a CNN network is incapable of handling different domains information from RGB and depth without prior knowledge about depth artifact.

Late fusion with probability map for RGB and depth is adopted in \cite{cheng2017localitysensitive} for segmentation, and here we generalize the network structure for normal estimation, by replacing the probability map with a binary mask indicating whether the depth pixel is available or not, giving the result in Fig.\,\ref{fig:fusion-loss-compare}(e). 
The role of binary mask we use is consistent with that of the probability map in \cite{cheng2017localitysensitive} which indicates how much the source is trustworthy. 
% Since valid depth pixels are mostly reliable, we use the binary mask for normal estimation. 
Similar to early fusion, the result of late fusion has noticeable artifacts along the depth holes indicating the fusion is not smooth. 

In light of this, single-scale fusion is not efficient for fusing RGB and depth when RGB and depth contain different noise. 
RGB is sensitive to lighting conditions while depth is corrupted at object edges and distant surfaces, indicating that the output from RGB and depth can be inconsistent. 
If depth is integrated into RGB in a single scale, the fusion is hard to eliminate the difference between two sources and give a smooth result.
% Further, the output at the end of RGB decoder depends merely on RGB input, and the large scale erroneous area cannot be rectified with a one-time fusion from depth decoder output, which 
This motivates us to merge depth features into RGB branch at four different scales in a hierarchical manner. 
In this way, the features from two branches are successively merged, where the global surface orientation error would be corrected at small resolution features, while detail refinement would take place at the final scale. 
As shown in Fig.\,\ref{fig:fusion-loss-compare}, the result of the proposed hierarchical fusion gives smoother result with detail well preserved.

\subsection{Confidence Map Estimation}
\label{sec:map}
% Motivation
While hierarchical fusion improves normal estimation over existing fusion schemes, further examination at pixels around depth holes shows that the transition is not smooth as shown in Fig.\,\ref{fig:mask-map-compare}(e) where the right side of the table has erroneous prediction close to depth hold boundary. 
This indicates that a binary masking is not sufficient for depth conditioning, and a more adaptive re-weighting would be more favorable. Therefore, a light-weight network for depth confidence map is designed as follows.

Depth along with a binary mask indicating missing pixels in depth are fed into a convolutional network with five layers as shown in Fig.\,\ref{fig:framework}, where the first two layers are with 3$\times$3 kernel size and the following three layers are with 1$\times$1 kernels. In this way, the receptive field is small enough to restrict local adaption to depth variation. 
Then the confidence map is down-sampled using shared pooling mask with depth branch and passed into the fusion module to facilitate fusion operation as described in Eq.\,\ref{eq:map}.
By comparing Fig.\,\ref{fig:mask-map-compare}(e) and (f), the confidence map leads to a more accurate fusion result, correcting the error at the right side of the table. 

To understand the role of the confidence map, we show the confidence map in Fig.\,\ref{fig:mask-map-compare}(d). The edge pixels are with the smallest confidence value indicating a high likelihood of outlier or noise, while the hole area is with a small yet non-zero value, suggesting that to enable smooth transition, information in depth holes can be passed into the merge result as long as RGB features take the dominant role.

\subsection{Hybrid Loss}
\label{hybrid-loss}

\deflen{realwidthm}{200pt}
\begin{figure}[t]
\centering
    \subfloat[RGB Image]{\includegraphics[width=\realwidthm]{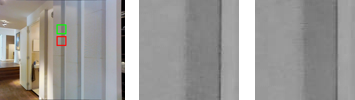}}\\
    \subfloat[Corresponding ground-truth normal]{\includegraphics[width=\realwidthm]{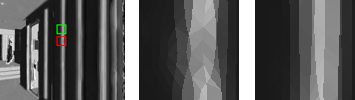}}
 \caption{Enlarged patches from input image and ground-truth normal map in horizontal direction. Upper row: input image, patch in red rectangle, patch in green rectangle. Bottom row: ground-truth normal map, patch in red rectangle, patch in green rectangle.}
 \label{fig:lossdesign}
\end{figure}

As mentioned in Section\,\ref{sec:intro}, we use Matterport3D and ScanNet datasets for training and testing because RGB-D data captured by camera and ground-truth normal pairs are provided. 
However, the ground-truth normal suffers from multiview reconstruction errors as shown in Fig.\,\ref{fig:lossdesign}(b) where the normal map is piece-wise constant inside the mesh triangular and the edge does not align with the RGB input. Given noisy ground-truth like this, improper handling of loss function during training will lead to deficient performance. The reason is as follows. 

Given the similar inputs in green and red rectangular in Fig.\,\ref{fig:lossdesign}(a), the output would be similar. However, the corresponding ground-truth normal maps are different as shown Fig.\,\ref{fig:lossdesign}(b), thus by minimizing the loss function, the network will learn an expectation of all pairs of input and ground-truth \cite{pmlr-v80-lehtinen18a}:
\begin{equation}\label{eq:loss_origin_multiple}
\min_{\theta} \quad \mathbb{E}_{( {\cal I}_c,{\cal I}_d,{\cal I}_n^{(\rm gt)} )} L({\cal I}_n^{(\rm gt)}, f_\theta({\cal I}_c, {\cal I}_d)).
\end{equation}
For $L_2$ loss $L_2({\cal I}_n^{(\rm gt)}, {\cal I}_n) = \| {\cal I}_n^{(\rm gt)} - {\cal I}_n\|_2^2$, the minimization will lead to an arithmetic mean of the observations, while $L_1$ loss $L_1({\cal I}_n^{(\rm gt)}, {\cal I}_n) = | {\cal I}_n^{(\rm gt)} - {\cal I}_n |$ will lead to median of the observations.

\begin{figure}[t]
\centering
    \subfloat[mean]{\includegraphics[width=\realwidthmm]{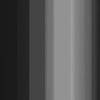}}\hspace{\gauinterm}
    \subfloat[median]{\includegraphics[width=\realwidthmm]{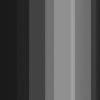}}
 \caption{Mean and median results from normal observations with the same RGB input.}
 \label{fig:lossdesign2}
\end{figure}

To see which loss is more proper for the given dataset, we sample patches along the edge in Fig.\,\ref{fig:lossdesign} with same horizontal position as patches in the color rectangles, and compute the mean and median normal results of these sampled patches shown in Fig.\,\ref{fig:lossdesign2} where both generate reasonable results though median result has sharper edges than mean result, indicating that $L_1$ loss will generate a more visually appealing result with sharp details. 

In this work, we adopt hybrid multi-scale loss function:
\begin{align}\label{eq:loss_hybird}
L({\cal I}_n^{(\rm gt)}, {\cal I}_n) = &\sum_{l=1,2} w_l L_2({\cal I}_n^{(\rm gt)}(l), {\cal I}_n(l)) \\ \nonumber
& \quad + \sum_{l=3,4} w_l L_1({\cal I}_n^{(\rm gt)}(l), {\cal I}_n(l)),
\end{align}
where $l=1,2,3,4$ denotes the scales from small to large, and $w_l$ is the weight for loss at different scales and is set to be $[0.2, 0.4, 0.8, 1.0]$. $L_1$ loss is used for large scale outputs for detail enhancement, while $L_2$ loss is used for coarse scale outputs for overall accuracy. 
Using hybrid loss generates clean and visually better result than $L_2$ loss widely used for normal estimation \cite{qi2018geonet, zhang2017physically,bansal2016marr} as shown in Fig.\,\ref{fig:fusion-loss-compare}. 
The proposed method is named as \textit{Hierarchical RGB-D Fusion with Confidence Map}, and referred to as \textit{HFM-Net} for short.

\deflen{gauinterm}{20pt}
\deflen{realwidthmm}{70pt}
%-------------------------------------------------------------------------
\section{Experiment}
\label{sec:exp}
\begin{table*}[t]
\begin{center}
  \begin{tabular}{c|c|cc|cc|cc|c}
  \hline
   \multicolumn{2}{c}{} &
   \multicolumn{2}{|c}{RGB-based} &
   \multicolumn{2}{|c}{Depth-based} &
   \multicolumn{2}{|c}{RGBD-based} & 
   \multicolumn{1}{|c}{Ours} \\
   \hline
   &         & Skip-Net & Zhang's & Levin's &  DC & GeoNet-D& GFMM&  HFM-Net\\
   & Metrics &         \cite{bansal2016marr}  & \cite{zhang2017physically} &    \cite{levin2004colorization}     &  \cite{zhang2018deep}              &      \cite{qi2018geonet} & \cite{Gong2013Guided}    & \\
   \hline
   \hline
   & mean               & 26.081  & 19.346 & 21.588   & 19.126   & 17.234    & 16.537    & \textbf{13.062}      \\
   Matter-& median       & 19.089  & 12.070 & 12.079   &  9.563   &  8.744    & 8.028     & \textbf{6.090}      \\
   port3D& 11.25$^\circ$  & 31.76   & 52.64  & 58.07    &  61.48   & 64.89     & 65.3      & \textbf{72.23}      \\
   & 22.5$^\circ$         & 57.61   & 72.12  & 69.59    & 74.08    & 78.5      & 79.94     & \textbf{84.41}      \\
   & 30$^\circ$           & 67.60   & 79.44  & 75.00    & 79.22    & 83.75     & 84.16     & \textbf{88.31}      \\
   \hline
   & mean                & 26.174   & 23.306  & 33.071   & 30.652   & 23.289    & 21.174     & \textbf{14.590}      \\
   Scan-& median          & 20.598   & 15.95  & 23.451   & 20.762   & 15.725    & 13.598     & \textbf{7.468}      \\
   Net& 11.25$^\circ$      & 28.78    & 40.43   & 34.52    & 39.35    & 46.41     & 50.78      & \textbf{65.65}      \\
   & 22.5$^\circ$          & 54.30    & 63.08   & 49.47    & 55.27    & 64.04     & 67.30      & \textbf{81.21}      \\ 
   & 30$^\circ$            & 67.00    & 71.88   & 56.37    & 60.03    & 76.78     & 77.00      & \textbf{86.21}      \\
   \hline
   & runtime             & 2.501s & 0.039s& 0.156+0.9s    & 0.156+0.058s    & 0.156+0.041s     & 0.156+0.041s      & 0.085s   \\ \hline
  \end{tabular}
 \end{center}
\caption{Performance of surface normal prediction on Matterport3D and ScanNet dataset.}
\label{tab:performance_in_sc}
\end{table*}   

\subsection{Implementation Details}
\deflen{gauinter}{0pt}
\deflen{gauwidth}{82pt}
\deflen{realwidth}{75pt}
\textbf{Dataset}
We evaluate our approach on two datasets, Matterport3D \cite{Matterport3D} and ScanNet \cite{dai2017scannet}. For the corresponding ground-truth normal data, we use the render normal provided by \cite{zhang2018deep} which was generated with multiview reconstruction.
% The 99 scenes of 
Matterport3D is divided into 105432 images for training and 11302 for testing; 
% 1307 scenes of 
ScanNet is divided into 59743 for training and 7517 for testing with file lists provided in \cite{zhang2018deep}. 
Since ground-truth normal data in the Matterport3D suffer from reconstruction noise, \eg, in outdoor scenes or mirror area, we remove the samples in the testing dataset with large error so as to avoid unreliable evaluation. 
After data pruning, 6.47\% (782 out of 12084) testing images are removed, leading to 11302 remaining. Details of data pruning can be found in the supplementary.

% For both dataset, in addition to depth, color and ground truth normal, we also generate the binarized depth mask image, represents whether the depth value obtained from the raw depth image at that position is exact. We set a threshold for filtering valid pixels.
% Similarly, we generate the normal mask from the normal image.

\textbf{Training Details}
We use RMSprop optimizer with initial learning rate set to $1e^{-3}$ and decayed at epoch $[2, 4, 6, 9, 12]$ with decay rate $0.5$. The model is trained from scratch without pretrained model for 15 epochs. 
We first use $L_2$ loss for all scales in the first 4 epochs and then change to hybrid loss defined in Eq.\,\ref{eq:loss_hybird} to ensure stable training at the beginning. 
We implement with PyTorch on NVIDIA GeForce GTX Titan X GPU.

\textbf{Evaluation Metrics}
The normal prediction performance is evaluated with five metrics. We compute the per-pixel angle distance between prediction and ground-truth, then compute mean and median for valid pixels with given ground-truth normal. 
% Since mean value is vulnerable to extreme values, median is used to make up for the inadequacy of mean in the skewed distribution.
In addition to mean and median, we also compute the fraction of pixels with angle difference with ground-truth less than $ t $ where $ t $ = 11.25$^\circ$, 22.5$^\circ$, and 30$^\circ$ as used in \cite{fouhey2013data}.

\subsection{Main Results}
We compare our proposed HFM-Net with the state-of-the-art normal estimation methods, which are classified into three categories in accordance with Section \ref{sec:related}, while normal-depth consistency based methods are adopted as alternatives for RGB-D fusion thus also put in the RGB-D category. 

\textbf{RGB-based} methods include Skip-Net \cite{bansal2016marr} and Zhang's algorithm \cite{zhang2017physically}. Pretrained models on Matterport3D and ScanNet of Zhang's are provided in \cite{zhang2018deep}, and Skip-Net is fine-tuned for Matterport3D and ScanNet based on the pre-trained model on NYUv2 dataset using public available training code.

\textbf{Depth-based}
Depth information is used to compute surface normal in existing works \cite{silberman2012indoor,dai2017scannet,Matterport3D} based on geometric relation between depth and surface normal.
Since the input depth is incomplete, we first implement depth inpainting before converting into normal map.
Two algorithms are used to preprocess the input depth images: colorization algorithm in \cite{levin2004colorization} (denoted as Levin's) as used in NYUv2 and the state-of-the-art depth completion (shortened as DC) \cite{zhang2018deep}.
% Many works design CNN for learning prediction, and geometric consistency can be also used for calculation.
After depth inpainting, we follow the same procedure in \cite{qi2018geonet} to generate normal from depth.

\textbf{RGBD-based}
For the RGB-D fusion methods, we adopt methods in GFMM \cite{Gong2013Guided} and the state-of-the-art GeoNet \cite{qi2018geonet} to merge depth input into initial RGB-based normal output for refinement.
Specifically, we choose Zhang's method \cite{zhang2017physically} for initial normal estimation from RGB, and calculate a rough normal from raw depth image at the same time, then merge the two normal estimations using methods in GFMM \cite{Gong2013Guided} and GeoNet \cite{qi2018geonet} to estimate the final surface normal map.

\deflen{realwidthx}{80pt}
\begin{figure*}[t]
\centering
 \subfloat[RGB Image]{\includegraphics[width=\realwidthx]{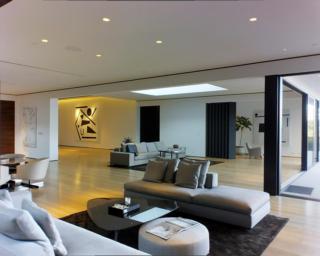}}\hspace{\gauinter}
 \subfloat[Depth Image]{\includegraphics[width=\realwidthx]{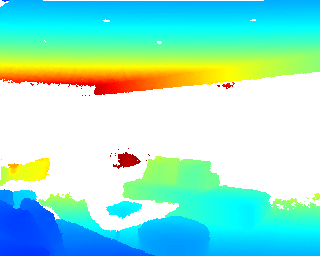}}\hspace{\gauinter}
 \subfloat[Ground-truth]{\includegraphics[width=\realwidthx]{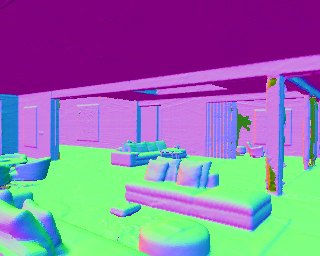}}\hspace{\gauinter}
 \subfloat[Skip-Net \cite{bansal2016marr}]{\includegraphics[width=\realwidthx]{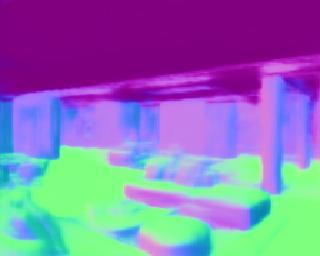}}\hspace{\gauinter}
  \subfloat[Zhang's \cite{zhang2017physically}]{\includegraphics[width=\realwidthx]{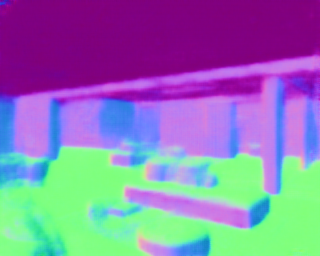}}\hspace{\gauinter}\\
 \subfloat[Levin's \cite{levin2004colorization} ]{\includegraphics[width=\realwidthx]{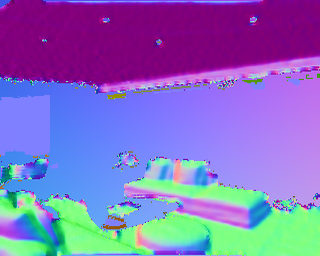}}\hspace{\gauinter}
 \subfloat[DC \cite{zhang2018deep}]{\includegraphics[width=\realwidthx]{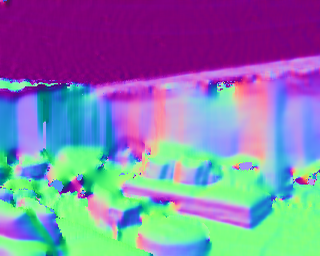}}\hspace{\gauinter}
 \subfloat[GeoNet-D \cite{qi2018geonet}]{\includegraphics[width=\realwidthx]{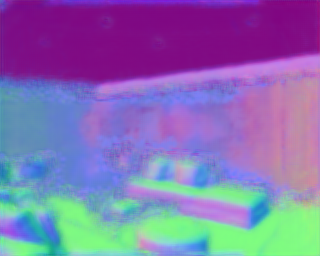}}\hspace{\gauinter}
 \subfloat[GFMM \cite{Liu2012Guided}]{\includegraphics[width=\realwidthx]{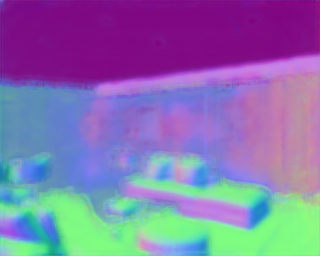}}\hspace{\gauinter}
 \subfloat[HFM-Net]{\includegraphics[width=\realwidthx]{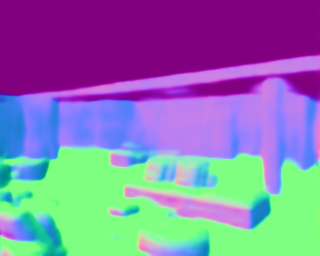}}\hspace{\gauinter}
 \caption{Surface normal estimation with different algorithms, test on Matterport3D dataset.} 
\label{all_model_matterport}
\end{figure*}

\begin{figure*}[t]
\centering
 \subfloat[RGB Image]{\includegraphics[width=\realwidthx]{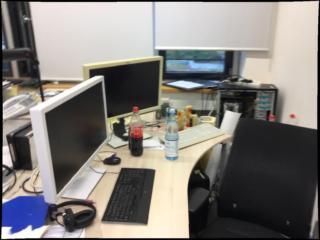}}\hspace{\gauinter}
 \subfloat[Depth Image]{\includegraphics[width=\realwidthx]{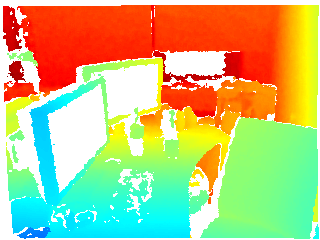}}\hspace{\gauinter}
 \subfloat[Ground-truth]{\includegraphics[width=\realwidthx]{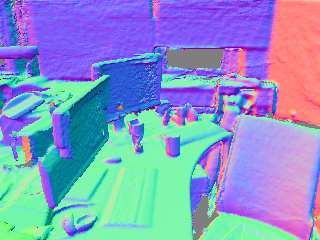}}\hspace{\gauinter}
 \subfloat[Skip-Net \cite{bansal2016marr}]{\includegraphics[width=\realwidthx]{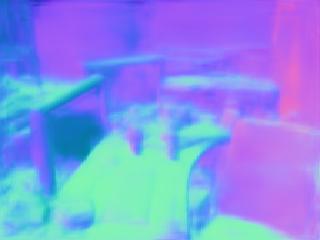}}\hspace{\gauinter}
  \subfloat[Zhang's \cite{zhang2017physically}]{\includegraphics[width=\realwidthx]{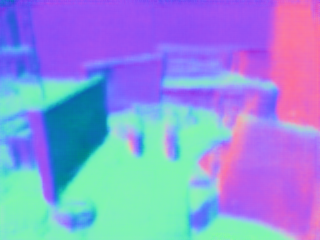}}\hspace{\gauinter}\\
 \subfloat[Levin's \cite{levin2004colorization} ]{\includegraphics[width=\realwidthx]{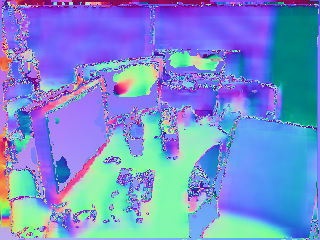}}\hspace{\gauinter}
 \subfloat[DC \cite{zhang2018deep}]{\includegraphics[width=\realwidthx]{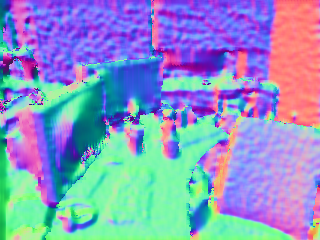}}\hspace{\gauinter}
 \subfloat[GeoNet-D \cite{qi2018geonet}]{\includegraphics[width=\realwidthx]{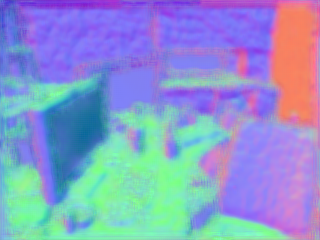}}\hspace{\gauinter}
 \subfloat[GFMM \cite{Liu2012Guided}]{\includegraphics[width=\realwidthx]{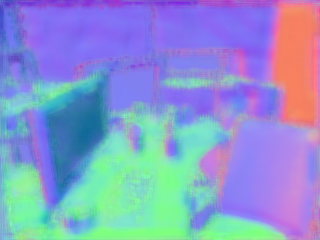}}\hspace{\gauinter}
 \subfloat[HFM-Net]{\includegraphics[width=\realwidthx]{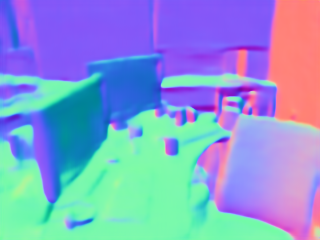}}\hspace{\gauinter}
 \caption{Surface normal estimation with different algorithms, test on ScanNet dataset.} 
\label{all_model_scannet}
\end{figure*}

We test on two datasets respectively with the five metrics as shown in Table \ref{tab:performance_in_sc}, where HFM-Net outperforms all the other schemes in different metrics. In terms of mean value, HFM-Net outperforms RGB-based methods by at least 6.284, and 6.064 over depth-inpainting based methods, and 3.475 over RGBD-based methods. 
Visual evaluation results are shown in Fig.\,\ref{all_model_matterport} and Fig.\,\ref{all_model_scannet}. 
RGB-based methods miss details such as the sofa in Fig.\,\ref{all_model_matterport} with blurry edges. Depth-based methods have serious errors at the depth hole regions and noticeable noise. Competing RGB-D fusion methods fail to generate accurate results at areas where depth is noisy or corrupted. On the contrary, our HFM-Net is exhibiting nice normal prediction both at smooth planar areas and along sharp edges.

\subsection{Ablation Study}
For better understanding of how HFM-Net works, we investigate the effect of each component in the network with the following ablation study.

\begin{figure}[h]
\centering
\subfloat[RGB Image]{\includegraphics[width=\realwidth]{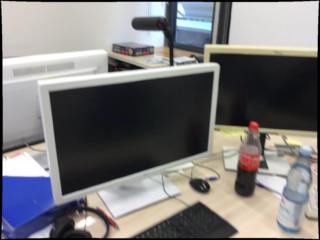}}\hspace{\gauinter}
\subfloat[Sensor Depth]{\includegraphics[width=\realwidth]{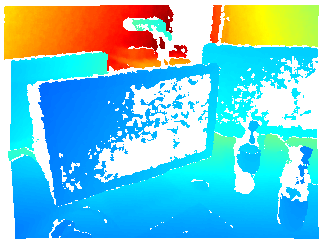}}\hspace{\gauinter}
\subfloat[Ground-truth]{\includegraphics[width=\realwidth]{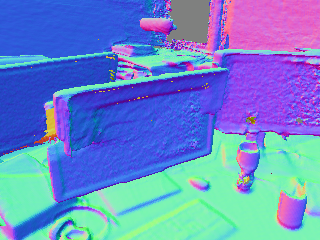}}\hspace{\gauinter}
\subfloat[Early Fusion]{\includegraphics[width=\realwidth]{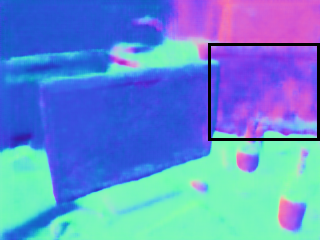}}\hspace{\gauinter}
\subfloat[Late Fusion]{\includegraphics[width=\realwidth]{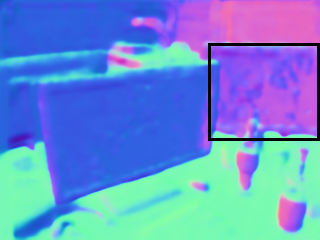}}\hspace{\gauinter}
\subfloat[Hierarchical Fusion]{\includegraphics[width=\realwidth]{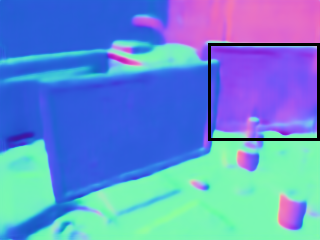}} \hspace{\gauinter}
 \subfloat[L2 Loss]{\includegraphics[width=\realwidth]{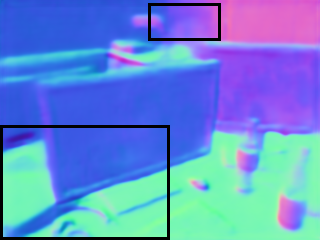}} \hspace{\gauinter}
  \subfloat[L1
  Loss]{\includegraphics[width=\realwidth]{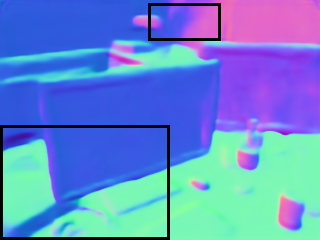}}\hspace{\gauinter}
  \subfloat[Hybrid loss]{\includegraphics[width=\realwidth]{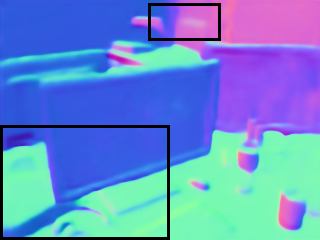}}\hspace{\gauinter}
  \subfloat[L2 Loss Detail]{\includegraphics[width=\realwidth]{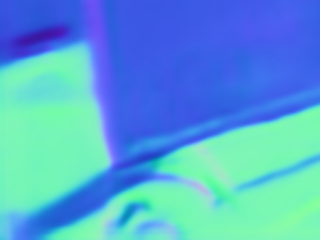}} \hspace{\gauinter}
  \subfloat[L1
  Loss Detail]{\includegraphics[width=\realwidth]{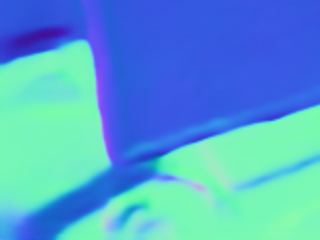}}\hspace{\gauinter}
  \subfloat[Hybrid Loss Detail]{\includegraphics[width=\realwidth]{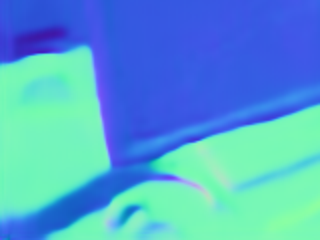}}\hspace{\gauinter}
 \caption{Surface normal estimation with different fusion schemes and different loss functions: (a) RGB input, (b) depth input, (c) ground truth, result of (d) early fusion, (e) late fusion, and (f) hierarchical fusion; result of using (g) $L_2$ loss, (h) $L_1$ loss, (i) proposed hybrid loss; (j-l) are the enlarged patches from (g-i). The hierarchical fusion produces a more accurate prediction in the area marked in black rectangles. 
 The hybrid loss design preserves the advantages of both $L_2$ (smooth surface) and $L_1$ loss (local details), with sharper details and more accurate results in depth holes.} 
 \label{fig:fusion-loss-compare}
\end{figure}

\textbf{Hierarchical Fusion} We compare hierarchical fusion (HF) with single-scale fusion including early fusion and late fusion as described Section \ref{sec:method}, denoted as Early-F and Late-F in Table \ref{tab:Ablation study} respectively. The binary mask is used for Late-F and HF, and all are trained using hybrid loss if not specified. As can be seen from Table \ref{tab:Ablation study}, Early-F and Late-F is less effective than HF+Mask+Hybrid, validating the use of HF. Furthermore, Fig.\,\ref{fig:fusion-loss-compare}(d-f) show the difference between single-scale and hierarchical fusion. The hierarchical fusion provides more accurate results in a planar surface especially in depth hole areas marked in black rectangles. 

\begin{figure}
\centering 
    \subfloat[RGB Image]{\includegraphics[width=\realwidth]{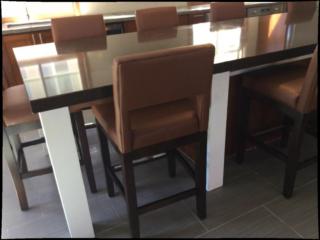}}\hspace{\gauinter}
    \subfloat[Sensor Depth]{\includegraphics[width=\realwidth]{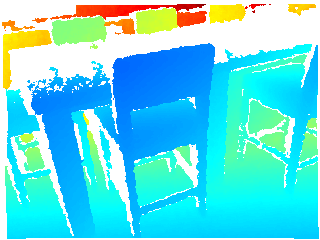}}\hspace{\gauinter}
    \subfloat[Ground-truth]{\includegraphics[width=\realwidth]{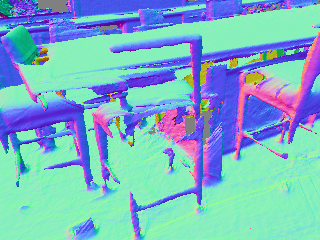}}\hspace{\gauinter} \subfloat[Confidence-map]{\includegraphics[width=\realwidth]{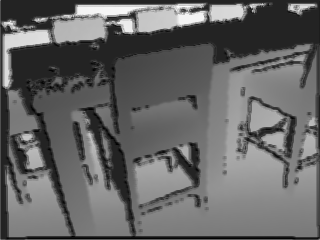}}\hspace{\gauinter}
    \subfloat[HF (Mask)]{\includegraphics[width=\realwidth]{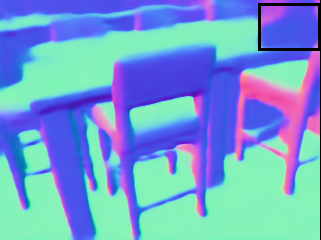}}\hspace{\gauinter}
     \subfloat[HF (Map)]{\includegraphics[width=\realwidth]{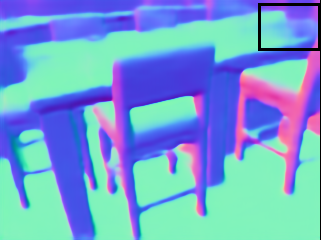}}\hspace{\gauinter}
     
    \caption{Surface normal estimation with different map/mask: (a) RGB input, (b) depth input, (c) ground truth, (d) confidence map, (e) hierarchical-fusion with mask, (f) hierarchical-fusion with map.} 
     \label{fig:mask-map-compare}
 \end{figure}

\textbf{Confidence Map}
We compare confidence map with binary mask. Fig.\,\ref{fig:mask-map-compare} shows the difference between fusion with confidence map and fusion with binary mask. Fusion with confidence map can reduce the negative effect of a depth hole during the fusion, and smooth the prediction around the boundary region of depth holes. 

\textbf{Hybrid Loss}
Apart from fusion method, different combinations of loss function are examined in the experiment. In comparison of hybrid loss, the confidence map is used in fusion. If the network use $L_2$ loss function in all layers, the prediction will tend to be blurry. On the other hand, a network with $L_1$ loss will tend to preserve more details. A hybrid loss function design, as described in Section \ref{hybrid-loss} can generate results with both smooth surface and fine object details, as shown in the comparison in Fig.\,\ref{fig:fusion-loss-compare} (g-l).

\begin{table}[h]
\begin{center}
 \scriptsize
  \begin{tabular}{c|c|c cccc}
   \hline
   &   & & & HF+Map & HF+Mask & HF+Map \\
      &  Metrics &  Early-F  & Late-F   &+L2   &+Hybrid   &+Hybrid \\
   \hline
   \hline
   & mean               & 13.968    &13.645  & 13.688   & 13.437   & \textbf{13.062}         \\
   Matter-& median       & 6.855     &6.567   & 7.235    &  6.507   &  \textbf{6.090}        \\
   port3D& 11.25$^\circ$  & 71.93     &70.79   & 69.21     &  70.98    & \textbf{72.23}         \\
   & 22.5$^\circ$         & 83.54     &83.68   & 83.45     & 83.96     &  \textbf{84.41}        \\
   & 30$^\circ$           & 87.44     &87.75   & 87.94     & 88.05     &  \textbf{88.31}        \\
   \hline
   & mean                & 16.045   & 17.425 & 14.946   & 14.696    &    \textbf{14.590}     \\
   Scan-& median          & 8.949    & 10.277 & 8.322    & 7.545     &    \textbf{7.468}       \\
   Net& 11.25$^\circ$      & 61.17    & 56.01  & 62.87     & 65.42    &  \textbf{65.65}        \\
   & 22.5$^\circ$          & 79.32    & 76.93  & 80.12     & 81.10    &  \textbf{81.21}        \\ 
   & 30$^\circ$            & 84.87    & 83.26  & 85.72     & 86.11    &  \textbf{86.21}      \\
   \hline
  \end{tabular}
\end{center}
\caption{Evaluation of variants of the proposed HFM-Net on Matterport3D and ScanNet datasets.}
\label{tab:Ablation study}
\end{table}

\subsection{Model Complexity and Runtime}
Table \ref{tab:performance_in_sc} reports the runtime of our method and other state-of-the-art methods. 
Skip-Net method uses the official evaluation code in MatCaffe. 
Levin's colorization method uses the code provided in NYUv2 dataset. 
GeoNet-D is the GeoNet with RGBD input, and we implement it in PyTorch. 
The consistency loss is added to GeoNet-D as a comparison scheme. 
The network forward runtime is averaged over Matterport3D test set with input images of size 320$\times$256 on NVIDIA GeForce GTX TITAN X GPU. 
Apart from the time cost in neural network forward pass, the runtime of depth-based and RGBD-based methods also includes the time spent on geometric calculation. 
As in shown in Table \ref{tab:performance_in_sc}, our method exceeds competing schemes in metric performance while taking a reasonably fast time.

%-------------------------------------------------------------------------
\section{Conclusion}
\label{sec:con}
In this work, we propose a hierarchical fusion scheme to combine RGB-D features at multiple scales with a confidence map estimated from depth input for depth conditioning to facilitate feature fusion.
Moreover, a hybrid loss function is designed to generate clean normal estimation even if the training targets suffer from reconstruction noise. 
Extensive experimental results demonstrate that our HFM-Net outperforms the state-of-the-art methods in providing more accurate surface normal prediction and sharper visually salient features.
Ablation studies validate the superiority of the proposed hierarchical fusion scheme over single-scale fusion schemes in existing works, the effectiveness of confidence map in producing accurate estimation around missing pixels in depth input, and the advantage of the hybrid loss function in overcoming dataset deficiency.

{\small
\bibliographystyle{ieee}
\bibliography{ms}

\begin{thebibliography}{10}\itemsep=-1pt

\bibitem{bansal2016marr}
A.~Bansal, B.~Russell, and A.~Gupta.
\newblock Marr revisited: 2d-3d alignment via surface normal prediction.
\newblock In {\em Proceedings of the IEEE Conference on Computer Vision and
  Pattern Recognition (CVPR)}, pages 5965--5974, 2016.

\bibitem{Matterport3D}
A.~Chang, A.~Dai, T.~Funkhouser, M.~Halber, M.~Niessner, M.~Savva, S.~Song,
  A.~Zeng, and Y.~Zhang.
\newblock {Matterport3D}: Learning from {RGB-D} data in indoor environments.
\newblock {\em International Conference on 3D Vision (3DV)}, 2017.

\bibitem{chen2017surface}
W.~Chen, D.~Xiang, and J.~Deng.
\newblock Surface normals in the wild.
\newblock In {\em Proceedings of the 2017 IEEE International Conference on
  Computer Vision, Venice, Italy}, pages 22--29, 2017.

\bibitem{cheng2017localitysensitive}
Y.~Cheng, R.~Cai, Z.~Li, X.~Zhao, and K.~Huang.
\newblock Locality sensitive deconvolution networks with gated fusion for rgb-d
  indoor semantic segmentation.
\newblock In {\em Proceedings of the IEEE Conference on Computer Vision and
  Pattern Recognition (CVPR)}, volume~3, 2017.

\bibitem{chu2018surfconv}
H.~Chu, W.-C. M.~K. Kundu, R.~Urtasun, and S.~Fidler.
\newblock Surfconv: Bridging 3d and 2d convolution for rgbd images.
\newblock In {\em Proceedings of the IEEE Conference on Computer Vision and
  Pattern Recognition (CVPR)}, pages 3002--3011, 2018.

\bibitem{dai2017scannet}
A.~Dai, A.~X. Chang, M.~Savva, M.~Halber, T.~Funkhouser, and M.~Nie{\ss}ner.
\newblock {ScanNet}: Richly-annotated 3d reconstructions of indoor scenes.
\newblock In {\em Proceedings of the IEEE Conference on Computer Vision and
  Pattern Recognition (CVPR)}, 2017.

\bibitem{Daniel2013Depth}
H.~C. Daniel, J.~Kannala, L.~Ladický, and J.~Heikkilä.
\newblock {\em Depth Map Inpainting under a Second-Order Smoothness Prior}.
\newblock Springer Berlin Heidelberg, 2013.

\bibitem{eigen2015predicting}
D.~Eigen and R.~Fergus.
\newblock Predicting depth, surface normals and semantic labels with a common
  multi-scale convolutional architecture.
\newblock In {\em Proceedings of the IEEE Conference on Computer Vision and
  Pattern Recognition (CVPR)}, pages 2650--2658, 2015.

\bibitem{fouhey2013data}
D.~F. Fouhey, A.~Gupta, and M.~Hebert.
\newblock Data-driven 3d primitives for single image understanding.
\newblock In {\em Proceedings of the IEEE Conference on Computer Vision and
  Pattern Recognition (CVPR)}, pages 3392--3399, 2013.

\bibitem{Gong2013Guided}
X.~Gong, J.~Liu, W.~Zhou, and J.~Liu.
\newblock Guided depth enhancement via a fast marching method ☆.
\newblock {\em Image \& Vision Computing}, 31(10):695--703, 2013.

\bibitem{gupta2014learning}
S.~Gupta, R.~Girshick, P.~Arbel{\'a}ez, and J.~Malik.
\newblock Learning rich features from rgb-d images for object detection and
  segmentation.
\newblock In {\em European Conference on Computer Vision (ECCV)}, pages
  345--360. Springer, 2014.

\bibitem{izadi2011kinectfusion}
S.~Izadi, D.~Kim, O.~Hilliges, D.~Molyneaux, R.~Newcombe, P.~Kohli, J.~Shotton,
  S.~Hodges, D.~Freeman, A.~Davison, et~al.
\newblock Kinectfusion: real-time 3d reconstruction and interaction using a
  moving depth camera.
\newblock In {\em Proceedings of the 24th annual ACM symposium on User
  interface software and technology}, pages 559--568. ACM, 2011.

\bibitem{pmlr-v80-lehtinen18a}
J.~Lehtinen, J.~Munkberg, J.~Hasselgren, S.~Laine, T.~Karras, M.~Aittala, and
  T.~Aila.
\newblock {N}oise2{N}oise: Learning image restoration without clean data.
\newblock In {\em Proceedings of the 35th International Conference on Machine
  Learning}, volume~80, pages 2965--2974, 2018.

\bibitem{levin2004colorization}
A.~Levin, D.~Lischinski, and Y.~Weiss.
\newblock Colorization using optimization.
\newblock In {\em ACM transactions on graphics ({TOG})}, volume~23, pages
  689--694. ACM, 2004.

\bibitem{Litany2018CVPR}
O.~Litany, A.~Bronstein, M.~Bronstein, and A.~Makadia.
\newblock Deformable shape completion with graph convolutional autoencoders.
\newblock In {\em Proceedings of the IEEE Conference on Computer Vision and
  Pattern Recognition (CVPR)}, pages 1886--1895, 2018.

\bibitem{Liu2012Guided}
J.~Liu, X.~Gong, and J.~Liu.
\newblock Guided inpainting and filtering for kinect depth maps.
\newblock In {\em International Conference on Pattern Recognition}, pages
  2055--2058, 2012.

\bibitem{long2015fully}
J.~Long, E.~Shelhamer, and T.~Darrell.
\newblock Fully convolutional networks for semantic segmentation.
\newblock In {\em Proceedings of the IEEE Conference on Computer Vision and
  Pattern Recognition (CVPR)}, pages 3431--3440, 2015.

\bibitem{newcombe2015dynamicfusion}
R.~A. Newcombe, D.~Fox, and S.~M. Seitz.
\newblock Dynamicfusion: Reconstruction and tracking of non-rigid scenes in
  real-time.
\newblock In {\em Proceedings of the IEEE conference on computer vision and
  pattern recognition}, pages 343--352, 2015.

\bibitem{Pang_2018_CVPR}
J.~Pang, W.~Sun, C.~Yang, J.~Ren, R.~Xiao, J.~Zeng, and L.~Lin.
\newblock Zoom and learn: Generalizing deep stereo matching to novel domains.
\newblock In {\em The IEEE Conference on Computer Vision and Pattern
  Recognition (CVPR)}, June 2018.

\bibitem{Qi2018CVPR}
C.~R. Qi, W.~Liu, C.~Wu, H.~Su, and L.~J. Guibas.
\newblock Frustum pointnets for 3d object detection from rgb-d data.
\newblock In {\em Proceedings of the IEEE Conference on Computer Vision and
  Pattern Recognition (CVPR)}, June 2018.

\bibitem{qi2018geonet}
X.~Qi, R.~Liao, Z.~Liu, R.~Urtasun, and J.~Jia.
\newblock Geonet: Geometric neural network for joint depth and surface normal
  estimation.
\newblock In {\em Proceedings of the IEEE Conference on Computer Vision and
  Pattern Recognition}, pages 283--291, 2018.

\bibitem{silberman2012indoor}
N.~Silberman, D.~Hoiem, P.~Kohli, and R.~Fergus.
\newblock Indoor segmentation and support inference from rgbd images.
\newblock In {\em European Conference on Computer Vision}, pages 746--760.
  Springer, 2012.

\bibitem{su2018splatnet}
H.~Su, V.~Jampani, D.~Sun, S.~Maji, E.~Kalogerakis, M.-H. Yang, and J.~Kautz.
\newblock Splatnet: Sparse lattice networks for point cloud processing.
\newblock In {\em Proceedings of the IEEE Conference on Computer Vision and
  Pattern Recognition (CVPR)}, pages 2530--2539, 2018.

\bibitem{Su_2018_CVPR}
S.~Su, F.~Heide, G.~Wetzstein, and W.~Heidrich.
\newblock Deep end-to-end time-of-flight imaging.
\newblock In {\em Proceedings of the IEEE Conference on Computer Vision and
  Pattern Recognition (CVPR)}, pages 6383--6392, 2018.

\bibitem{Thabet20143D}
A.~K. Thabet, J.~Lahoud, D.~Asmar, and B.~Ghanem.
\newblock 3d aware correction and completion of depth maps in piecewise planar
  scenes.
\newblock In {\em Asian Conference on Computer Vision}, pages 226--241, 2014.

\bibitem{wang2016surge}
P.~Wang, X.~Shen, B.~Russell, S.~Cohen, B.~Price, and A.~L. Yuille.
\newblock Surge: Surface regularized geometry estimation from a single image.
\newblock In {\em Advances in Neural Information Processing Systems}, pages
  172--180, 2016.

\bibitem{wang2018depthconv}
W.~Wang and U.~Neumann.
\newblock Depth-aware cnn for rgb-d segmentation.
\newblock In {\em European Conference on Computer Vision (ECCV)}. Springer,
  2018.

\bibitem{wang2015designing}
X.~Wang, D.~Fouhey, and A.~Gupta.
\newblock Designing deep networks for surface normal estimation.
\newblock In {\em Proceedings of the IEEE Conference on Computer Vision and
  Pattern Recognition (CVPR)}, pages 539--547, 2015.

\bibitem{zeng20183d}
J.~Zeng, G.~Cheung, M.~Ng, J.~Pang, and C.~Yang.
\newblock 3d point cloud denoising using graph laplacian regularization of a
  low dimensional manifold model.
\newblock {\em arXiv preprint arXiv:1803.07252}, 2018.

\bibitem{zeng2018deep}
J.~Zeng, J.~Pang, W.~Sun, G.~Cheung, and R.~Xiao.
\newblock Deep graph laplacian regularization.
\newblock {\em arXiv preprint arXiv:1807.11637}, 2018.

\bibitem{Zhang2017Probability}
H.-T. Zhang, J.~Yu, and Z.-F. Wang.
\newblock Probability contour guided depth map inpainting and superresolution
  using non-local total generalized variation.
\newblock {\em Multimedia Tools and Applications}, 77(7):9003--9020, 2018.

\bibitem{zhang2018deep}
Y.~Zhang and T.~Funkhouser.
\newblock Deep depth completion of a single rgb-d image.
\newblock In {\em Proceedings of the IEEE Conference on Computer Vision and
  Pattern Recognition (CVPR)}, pages 175--185, 2018.

\bibitem{zhang2017physically}
Y.~Zhang, S.~Song, E.~Yumer, M.~Savva, J.-Y. Lee, H.~Jin, and T.~Funkhouser.
\newblock Physically-based rendering for indoor scene understanding using
  convolutional neural networks.
\newblock In {\em Proceedings of the IEEE Conference on Computer Vision and
  Pattern Recognition (CVPR)}, pages 5057--5065. IEEE, 2017.

\end{thebibliography}
}

\end{document}